\documentclass[10pt,twocolumn,letterpaper]{article}

\usepackage[pagenumbers]{wacv}

\usepackage{graphicx}
\usepackage{amsmath}
\usepackage{amssymb}
\usepackage{booktabs}
\usepackage{comment}
\usepackage[pagebackref,breaklinks,colorlinks]{hyperref}

\usepackage[capitalize]{cleveref}
\crefname{section}{Sec.}{Secs.}
\Crefname{section}{Section}{Sections}
\Crefname{table}{Table}{Tables}
\crefname{table}{Tab.}{Tabs.}

\begin{document}

\title{Learning with Multi-modal Gradient Attention \\ for Explainable Composed Image Retrieval}
\author{Prateksha Udhayanan, Srikrishna Karanam, and Balaji Vasan Srinivasan\\
Adobe Research, Bangalore India \\
{\tt \scalebox{.7}{\{udhayana,skaranam,balsrini\}@adobe.com,}}
}
\newcommand{\pnote}[1]{\textcolor{red}{\textbf{PU: #1}}}

\maketitle

\begin{abstract}
We consider the problem of composed image retrieval that takes an input query consisting of an image and a modification text indicating the desired changes to be made on the image and retrieves images that match these changes. Current state-of-the-art techniques that address this problem use global features for the retrieval, resulting in incorrect localization of the regions of interest to be modified because of the global nature of the features, more so in cases of real-world, in-the-wild images. Since modifier texts usually correspond to specific local changes in an image, it is critical that models learn local features to be able to both localize and retrieve better. To this end, our key novelty is a new gradient-attention-based learning objective that explicitly forces the model to focus on the local regions of interest being modified in each retrieval step. We achieve this by first proposing a new visual image attention computation technique, which we call multi-modal gradient attention (MMGrad) that is explicitly conditioned on the modifier text. We next demonstrate how MMGrad can be incorporated into an end-to-end model training strategy with a new learning objective that explicitly forces these MMGrad attention maps to highlight the correct local regions corresponding to the modifier text. By training retrieval models with this new loss function, we show improved grounding by means of better visual attention maps, leading to better explainability of the models as well as competitive quantitative retrieval performance on standard benchmark datasets.
\end{abstract}

\section{Introduction}
\label{sec:intro}

\begin{figure*}
    \centering
    \includegraphics[width=0.85\linewidth]{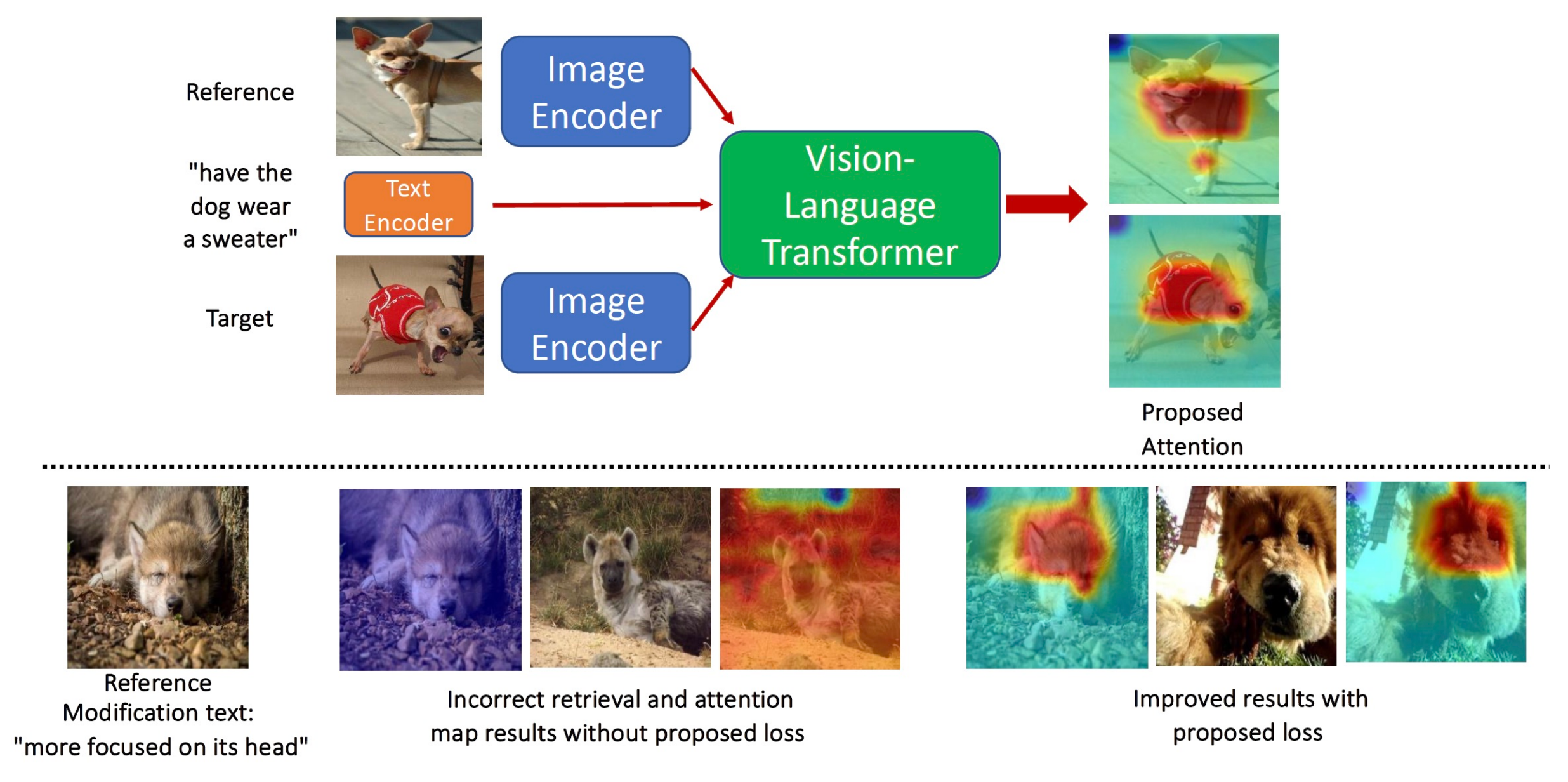}
    \caption{Top: Our proposed multi-modal gradient attention (MMGrad) approach that localizes the attention to specific regions of the target image guided by the modification text. Bottom: Results with and without the MMGrad loss illustrating improved performance.}
    \label{fig:teaser}
\end{figure*}

We consider the problem of \textit{composed image retrieval} where given an image and a modification text, our task is to retrieve images from a database that matches the input image, with the modifications in the text incorporated. For example, in Figure~\ref{fig:teaser}, for an image of a dog (``Reference'') and the modification text \textit{have the dog wear a sweater}, the expected retrieved image will be one of a dog with a sweater (``Target''). While the traditional image matching setup \cite{zheng2017sift} considers only image data and matches the semantics of the retrieved images, composed-image-retrieval task adds the text modality as well, necessitating the learning of a joint image-text feature space. There has been much recent work~\cite{maaf, mrn, tirg} that tackles this problem for a specific domain of data, e.g., the fashion domain with the Fashion-IQ \cite{fashioniq} dataset. This dataset focuses on specific types of attribute changes, e.g., change in color, length, shape, etc. Some example modifier texts in these cases are: \textit{has a blue print and short sleeves}, \textit{change red to green}, and \textit{change mesh to sequin}. These modifier texts can be categorized into a set of attribute changes and describe simple modifications. On the other hand, our focus is primarily on addressing the problem in the context of real-life in-the-wild images that depict both complex scenarios and convey fine-grained details through the modifier texts.

Existing techniques that address this problem for in-the-wild data train models that leverage global image and text features to learn the joint space. For instance, recent works \cite{cirplant,goenka2022fashionvlp,chen2022composed} directly use representations from standard image and text encoders before training image-text transformers that produce the ``composed'' features used for matching. Such an approach does not explicitly model locality, a unique aspect of our problem at hand. Our intuition is simple- as can be seen from Figure~\ref{fig:teaser} (top), since the objects of interest in the query occupy specific local image regions, the features used for image matching must ``focus" on these regions. This necessitates a learning scheme that explicitly forces the model to produce such features. Consequently, the key question we ask and address in this work is- \textit{how can we train models that highlight and focus on the modifier regions in images during the learning process?}

We address the aforementioned challenges with a new learning approach that builds on top of gradient-based attention to force models to focus on the regions of interest. With gradient attention, our key insight is to generate and refine attention maps during the learning process itself. For example, see the image in Figure~\ref{fig:teaser} (top) where the model's gradient attention is focused on the local region of interest in the image (the sweater region). To realize this, we propose a new loss function that makes such attention maps a principled part of the learning process by forcing them to be close to the local regions specified in the modifier text. 

In Figure~\ref{fig:teaser} (bottom), we show some example results with and without this proposed loss function. One can note that without our proposed loss, retrieved top-1 image is incorrect since the features are not localized (see the attention map). On the other hand, with our proposed loss, the attention map is localized in the region of interest, leading to the retrieval of the correct target image as the rank-1 result. While all prior work in attention learning has focused solely on image-only models \cite{selvaraju2017grad,chen2020adapting,ijcai2022p241}, our novelty lies in designing a new way to compute image attention maps conditioned on text input, leading to a new attention computation technique we call multi-modal gradient attention (MMGrad).  With all operations in the computation of MMGrad attention maps relying only on differentiable functions, we show one can train models with an explicit loss on the MMGrad attention maps in an end-to-end fashion, resulting in model attention focused on local regions of interest. Note that training models with such an explicit loss on these multi-modal attention maps is a key contribution of our paper. Consequently, the features produced by the model will have been more local in nature, which we show results in improvements in downstream retrieval performance when compared to the current state-of-the-art methods that use global features. Furthermore, with our method, we also obtain attention maps that highlight the regions of change as a byproduct, leading to a step towards model explainability for composed image retrieval.Note that the proposed loss function and training methodology can in principle be integrated with any backbone for this problem to enhance both baseline quantitative performance as well as inject visual explainability to its outputs.

To summarize, our key contributions are listed below:

\begin{itemize}
\item We propose a new multi-modal gradient attention computation mechanism for composed image retrieval that computes gradient attention maps conditioned on both input images as well as the modification text. Our method is flexible to be plugged into any backbone architecture for this problem. 
\item We propose an end-to-end learning scheme for vision-language Siamese transformers that integrates our proposed multi-modal gradient attention, leading to the models to focus on and generate more localized image representations suitable for retrieval tasks.
\item We conduct extensive experiments on standard benchmark datasets and demonstrate the efficacy and impact of our methods both qualitatively by means of attention maps and quantitatively with standard metrics. 
\end{itemize}

\begin{figure*}
    \centering
    \includegraphics[width=0.96\linewidth]{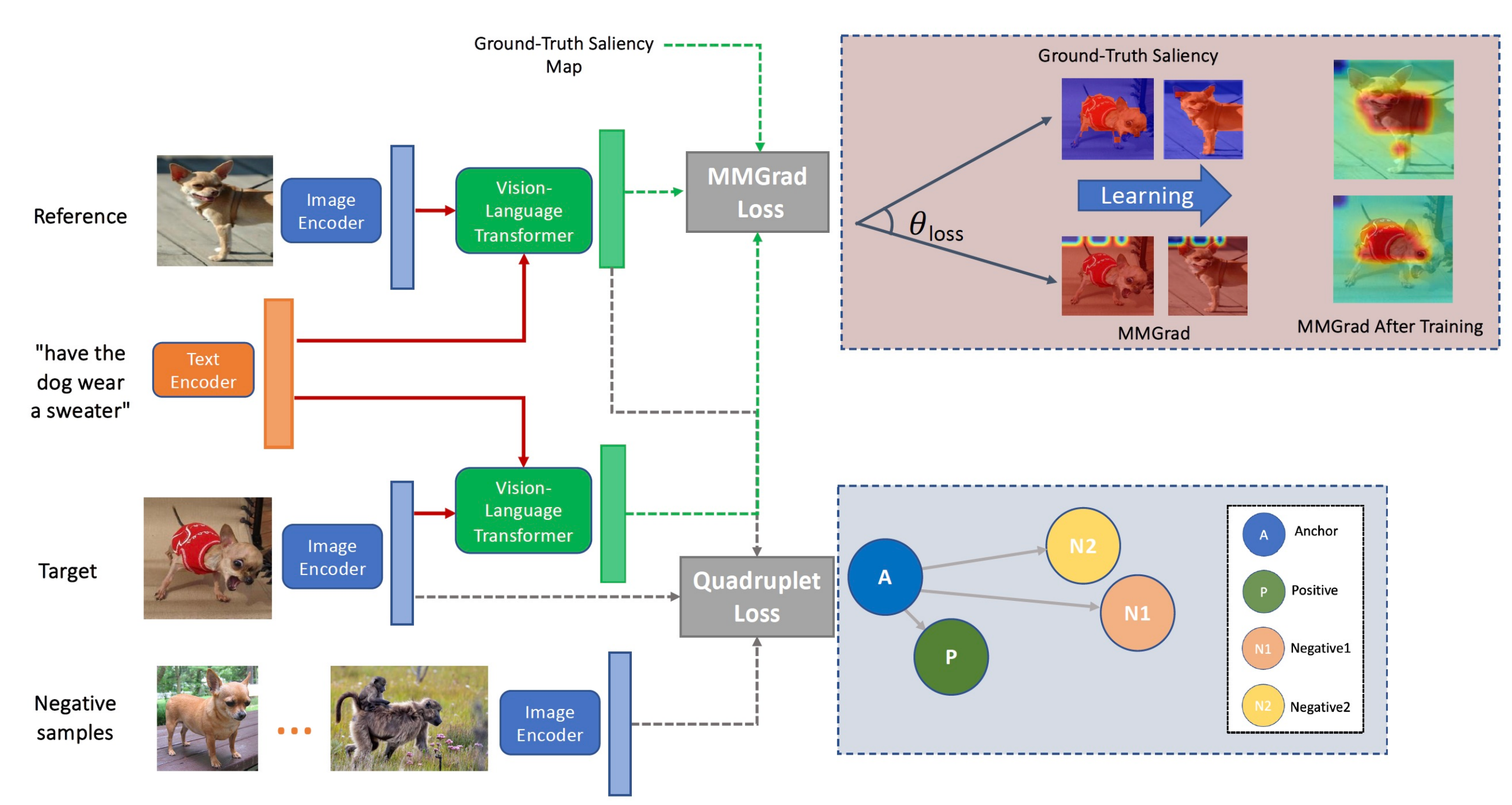}
    \caption{The architecture of our pipeline to train end-to-end Multi-modal transformer models with our proposed MMGrad attention and the resulting attention learning objective.}
    \label{fig:pipeline}
\end{figure*}

\section{Related work}
While there is much prior work in image retrieval, composed image retrieval is a relatively newer problem. Here, we briefly review relevant existing methods. \\ 
\textbf{Image Retrieval}: Image retrieval is a very well established problem in computer vision \cite{rui1999image,datta2008image,zheng2017sift} with significant progress to the point of being deployed at scale in production in the industry \cite{zhai2017visual,yang2017visual,zhang2018visual}. These problems are typically based on image-only queries, e.g., face recognition \cite{facerecognition} and visual product search \cite{liu2016deepfashion,zhang2018visual}, or cross-modal queries such as text-to-image \cite{radford2021learning} or sketch-to-image retrieval \cite{bhunia2022sketching}. \textit{Composed image retrieval}, on the other hand, is a slightly different problem where we are to retrieve an image that jointly satisfies an image and a modification text. \\
\textbf{Composed Image Retrieval}: There has been much recent work in this area. Vo et al.~\cite{tirg} proposed a gated-residual connection mechanism to fuse and jointly learn image and text features.  In Dodds et al. \cite{maaf}, a modality-agnostic fusion strategy was proposed where convolutional image features and the text embeddings were treated as modality-agnostic tokens for a multi-modal transformer model. In Anwaar et al. \cite{composeae}, an autoencoder was trained to compose image and text features using a metric learning strategy. In Cirplant \cite{cirplant}, a vision-language transformer built on top of pretrained models was proposed that used image features from ImageNet-trained convolutional models as part of a triplet loss learning framework. More recent works like ARTEMIS~\cite{artemis} used an MLP with attention to select the important characteristics of the target image. While these methods report impressive progress in this area, they learn and use global image representations for retrieval. However, since the problem is directly local in nature (local image regions to be modified based on the input text), our key difference and novelty is to (a) highlight the local regions that are being modified with a new multi-modal gradient (MMGrad) attention operation, and (b) train models end-to-end by integrating MMGrad in a new attention-based loss.\\
\textbf{Visual Attention Learning}: There has been much progress in gradient-based attention learning since the methods of Selvaraju et al. \cite{selvaraju2017grad} and Zhou et al. \cite{zhou2016learning}. While these methods were specifically designed for classification tasks, extensions to other tasks and models such as metric learning \cite{ijcai2022p241,chen2020adapting} and generative models \cite{liu2020towards} have also appeared in subsequently published papers. Further, there has also been progress in improving GradCAM attention maps \cite{selvaraju2017grad,chattopadhay2018grad,jiang2021layercam}. Our work is substantially different from these image-only methods in that we present a new technique to compute image attention maps conditioned on text input, leading to multi-modal gradient attention that helps highlight the local regions of change corresponding to a modification text.

\begin{figure*}
    \centering
    \includegraphics[width=0.75\linewidth]{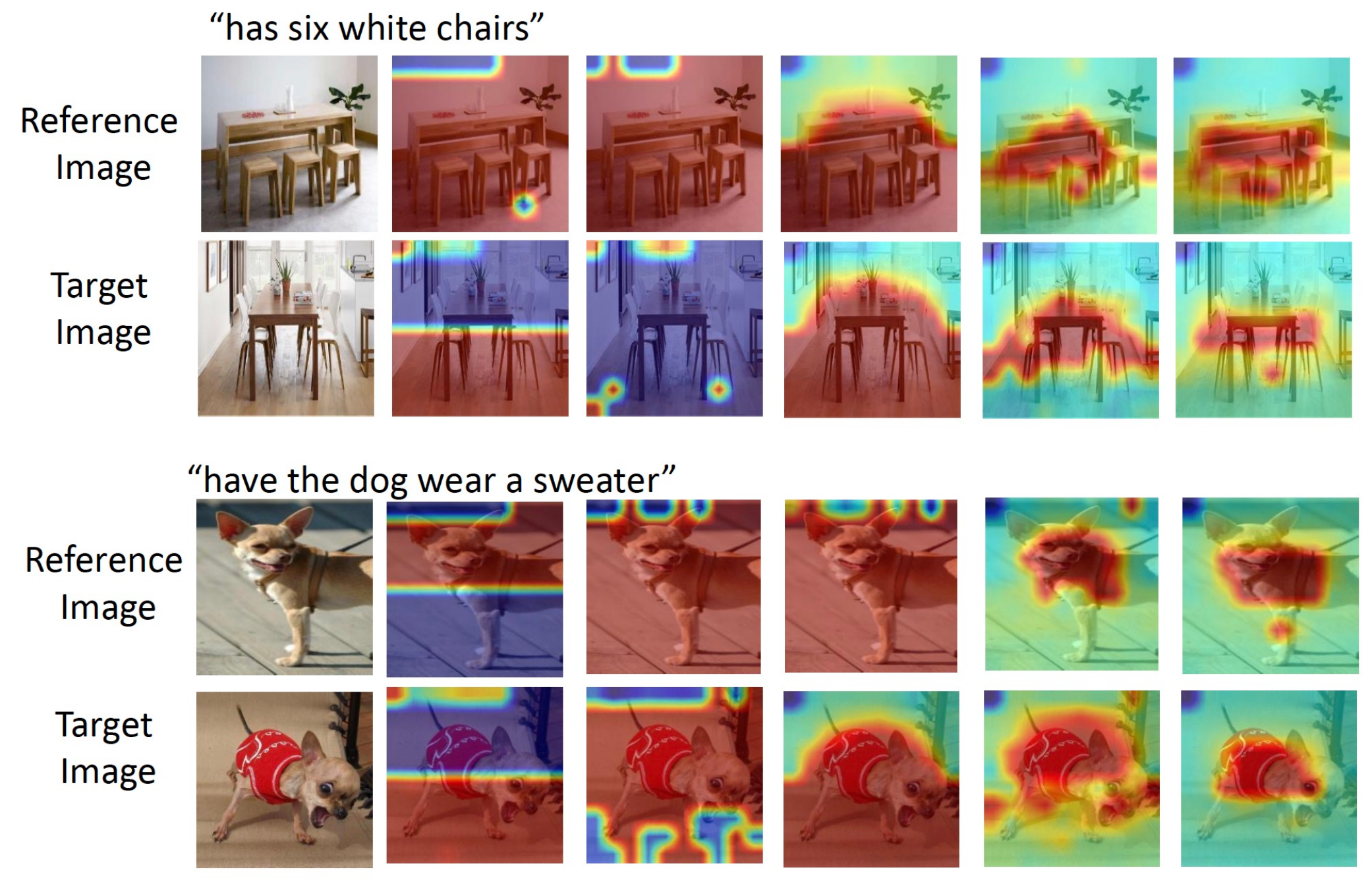}
    \caption{With our proposed MMGrad attention loss, the localization to the correct local modification region in our attention maps improve with training epochs, eventually getting close to the ground-truth saliency maps.}
    \label{fig:attentionEvolution}
\end{figure*}

\section{Proposed Approach}
We are interested in the problem of composed image retrieval where we have a reference image $\mathbf{I}_\text{ref}$, a modification text $\mathbf{T}_\text{mod}$, and a corresponding target image $\mathbf{I}_\text{tgt}$ that satisfies the modification of $\mathbf{I}_\text{ref}$ according to $\mathbf{T}_\text{mod}$. We propose to train Siamese models that push the jointly-learned feature vector of $\mathbf{I}_\text{ref}$ and $\mathbf{T}_\text{mod}$ close to that of $\mathbf{I}_\text{tgt}$. While existing methods such as Cirplant \cite{cirplant} train models to produce global feature representations, a key challenge with the problem setup is its local nature. Since the change between the reference and target image is typically local in nature (e.g., change in some part of the image is emphasized in $\mathbf{T}_\text{mod}$, e.g., see the sweater region in Fig~\ref{fig:teaser} (top)), one would want models to produce features that are influenced by these local regions. For instance, looking at the result in Figure~\ref{fig:teaser} (bottom) again, in the absence of such a local training strategy, the attention map of models trained in a global fashion can highlight a lot of background/distractor regions (e.g., the red regions surronding the dog image in the incorrect-results part of this figure). To achieve such focus on local regions during training, our key novelty is a new gradient attention mechanism we call multi-modal gradient attention (MMGrad) which highlights, by means of attention maps, the local regions in the images that are being modified. Further, to ensure the model actually learns local features to be able to produce accurate MMGrad maps, we also propose a new learning objective that shows how the MMGrad attention map can be integrated into an end-to-end Siamese training strategy. Our proposed model architecture is visually illustrated in Figure~\ref{fig:pipeline} which we next discuss in detail. 

Let the image encoder (realized with a vision transformer \cite{vit}) shown in Figure~\ref{fig:pipeline} be represented by $\mathbf{E}_\text{img}$. Note that our model is Siamese in nature with weight sharing across both the reference and target branches. Let the joint image-text model (realized with a vision-language transformer \cite{visualbert}) be $\mathbf{E}_\text{img-text}$. Given $\mathbf{I}_\text{ref}$ and $\mathbf{I}_\text{tgt}$, we first use $\mathbf{E}_\text{img}$ to obtain their feature vectors $\mathbf{f}_\text{ref}$ and $\mathbf{f}_\text{tgt}$. We also tokenize the modifier text $\mathbf{T}_\text{mod}$ to obtain the tokens $\mathbf{t}_\text{mod}$. Given $\mathbf{f}_\text{ref}$ and $\mathbf{t}_\text{mod}$, we use $\mathbf{E}_\text{img-text}$ to get the joint feature vector  $\mathbf{f}_\text{ref-mod}$. Similarly, we get the joint feature vector $\mathbf{f}_\text{tgt-mod}$ for the target image feature vector $\mathbf{f}_\text{tgt}$ and $\mathbf{t}_\text{mod}$. 

In the following section, we explain how we use these feature vectors that are explicitly conditioned on the text input to compute our proposed multi-modal gradient (MMGrad) attention maps that highlight local image regions that will be modified during the composed image retrieval step.

 \begin{figure*}[th]
    \centering
    \includegraphics[width=0.8\linewidth]{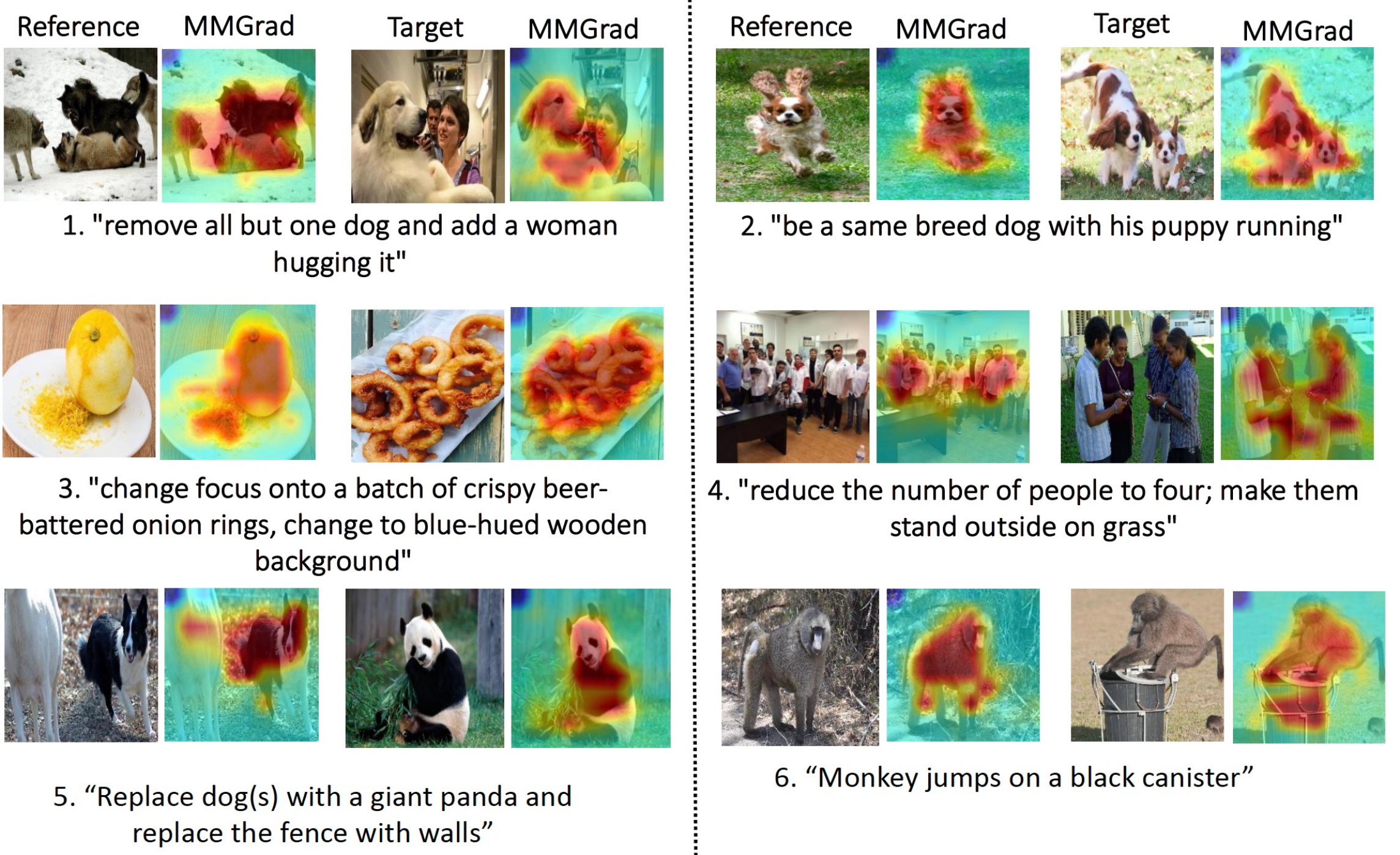}
    \caption{Qualitative results on validation data from CIRR with our proposed method where we see correct target images retrieved as rank-1 matches and MMGrad attention maps highlighting the right local regions corresponding to the modification text.}
    \label{fig:qualResults}
\end{figure*}

\subsection{MMGrad: Multi-modal Gradient Attention}
\label{sec:mmgrad}

As discussed above and in Section~\ref{sec:intro}, we seek to highlight local regions in images where there are modifications based on the modification text. To compute such an attention map, we propose to use the text-conditioned feature vectors to get the gradients necessary for the attention map. Our intuition is since these feature vectors are informed by the modification text, using them to compute the attention map and training models with a loss on this map (discussed below) will drive the model towards producing feature vectors that will be influenced by these local regions.

Given the joint feature vector for the target image $\mathbf{f}_\text{tgt-mod}$, we first obtain the gradients from the last layer of the $\mathbf{E}_\text{img-text}$ module. To do this, we will need a scalar score which we differentiate with respect to the parameters of the last layer of $\mathbf{E}_\text{img-text}$. We realize this by considering both reference image and target image joint feature vectors $\mathbf{f}_\text{ref-mod}$ and $\mathbf{f}_\text{tgt-mod}$, and compute their cosine similarity

\begin{equation}
    s=\dfrac {\mathbf{f}_\text{ref-mod} \cdot \mathbf{f}_\text{tgt-mod}} {\left\| \mathbf{f}_\text{ref-mod}\right\| _{2}\left\| \mathbf{f}_\text{tgt-mod}\right\| _{2}}
\end{equation}

Given $s$, we compute its derivatives with respect to the last dense (fully-connected) layer $\mathbf{L}$ of $\mathbf{E}_\text{img-text}$ to obtain a vector of gradients $\mathbf{g}=\frac{\partial s}{\partial \mathbf{L}}$. To get the attention map $\mathbf{M}$ on the target image, we first reshape its joint feature vector as $\mathbf{f}_\text{tgt-mod} \in \mathcal{R}^{d \times m \times n}$, where $d$ is the feature dimensionality (that also happens to be the number of neurons in the last layer of $\mathbf{E}_\text{img-text}$) and $m \times n$ is the spatial size of the attention map. $\mathbf{M}$ is then computed as:

\begin{equation}
\mathbf{M}=\text{ReLU}(\sum_{i=1}^{d}g_{i}\mathbf{f}_\text{tgt-mod}^{i})
\end{equation}

where each $\mathbf{f}_\text{tgt-mod}^{i} \in \mathcal{R}^{m\times n}$ and $g_i$ refers to the importance of neuron $i$ of the last layer of $E_{img-text}$ in retrieving the target image. 

Our idea with the proposed method above is to capture, in the features and hence the resulting attention map, the differences in the target image when compared to the reference image conditioned on the input modifier text. Note also that this is substantially different from existing work in attention learning \cite{selvaraju2017grad,chen2020adapting,ijcai2022p241} since these methods focus solely on image-only models whereas our attention map is multimodal and explicitly conditioned on a jointly learned image-text feature vector output from an image-text transformer model.

\begin{figure*}[th]
    \centering
    \includegraphics[width=0.9\linewidth]{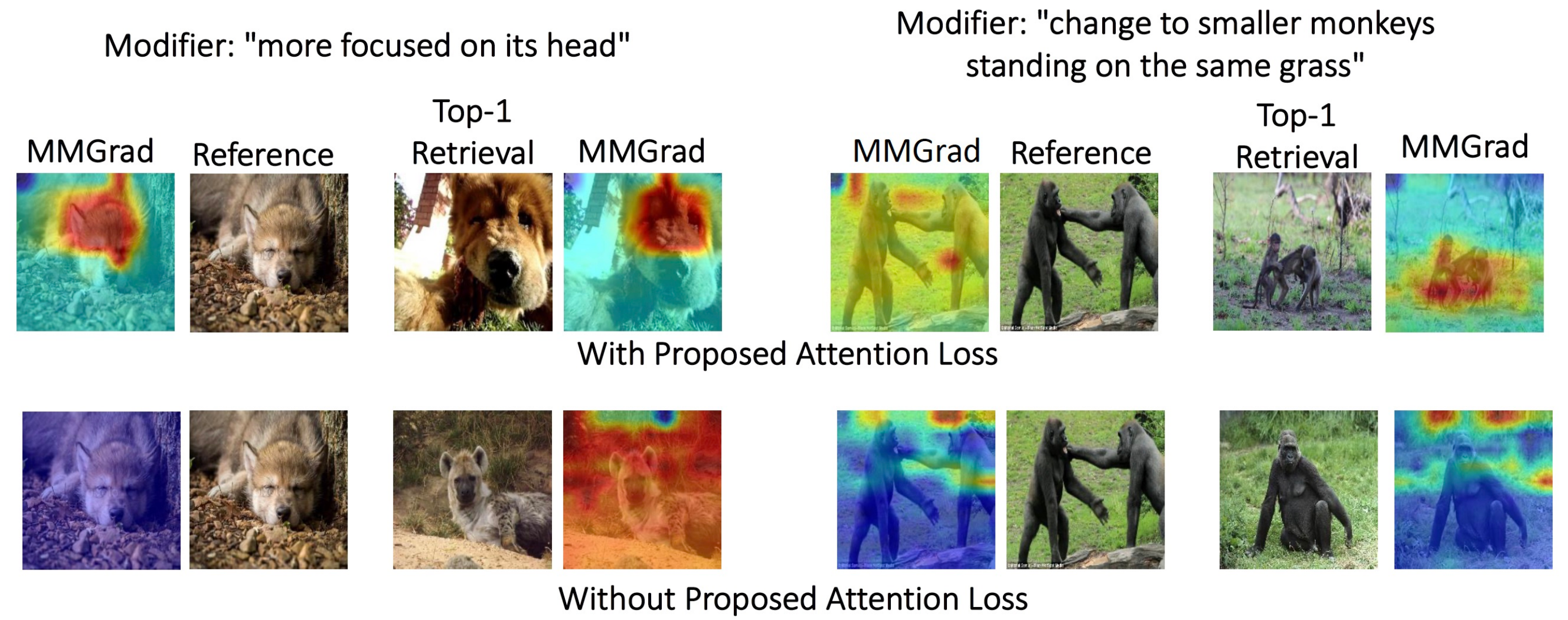}
    \caption{Qualitative results of our model trained with and without the proposed MMGrad attention loss. One can note our attention loss leads to improved rank-1 retrievals as well as attention maps that highlight the correct regions corresponding to the modification text.}
    \label{fig:withWithoutAttention}
\end{figure*}

\subsection{Learning with MMGrad Attention}
During the model training phase, we want our MMGrad attention map $\mathbf{M}$ to be as close as possible to the local region corresponding to the modifier text in the input. To be able to do this, we first generate ground truth saliency maps for all training samples. We use an off-the-shelf text-prompt-conditioned image segmentation model \cite{clipseg} for this purpose. We extract key phrases \cite{rake} from the modifier text, use them as prompt inputs to the model, and obtain the saliency maps corresponding to each key phrase. Finally, we aggregate all these maps, binarize the result, and use it as the ground truth saliency map $\mathbf{S}$. 

For an input sample ($\mathbf{I}_\text{ref}$, $\mathbf{T}_\text{mod}$, and $\mathbf{I}_\text{tgt}$), we compute our multi-modal gradient attention map $\mathbf{M}$ as discussed in Section~\ref{sec:mmgrad} and seek to make it close to the corresponding saliency map $\mathbf{S}$. We achieve this with a learning objective based on cosine distance as:

\begin{equation}
    \mathcal{L}_\text{MMGrad} = 1 - \dfrac{\langle\mathbf{M} , \mathbf{S}\rangle}{||\mathbf{M}||||\mathbf{S}||}
    \label{eq:mmgradLoss}
\end{equation}

where $\langle\mathbf{M} , \mathbf{S}\rangle$ represents the inner product between the flattened versions of $\mathbf{M}$ and $\mathbf{S}$, and $||\mathbf{M}||$, $||\mathbf{S}||$ are their corresponding Euclidean norms. On training with Equation~\ref{eq:mmgradLoss}, our MMGrad attention map $\mathbf{M}$ gets closer to ground-truth saliency $\textbf{S}$, leading to the model to highlight the modified local regions on the image. To concretely show the impact of $\mathcal{L}_\text{MMGrad}$, we provide qualitative results in Figure~\ref{fig:attentionEvolution}, where one can note two cases (one in each row). From the third column onwards, we show the progression of $\mathbf{M}$ with training epochs. At the earlier stages of training, $\mathbf{M}$ does not attend to the correct regions, and as training progresses, this improves, with the final $\mathbf{M}$ close to $\textbf{S}$.

\subsection{Overall Loss Function}
In addition to the proposed multi-modal gradient attention loss of Equation~\ref{eq:mmgradLoss}, we also use a standard metric learning loss to ensure a discriminative feature space. In particular, we use the quadruplet loss:

\begin{equation}
\begin{split}
    \mathcal{L}_\text{quad} & = max(d_1^2-d_2^2+m_1, 0) \\
    & + max(d_1^2 - d_3^2 + m_2, 0)
    \label{eq:gradVLLoss}
\end{split}
\end{equation}
where $d_1$ is the distance between the anchor and the positive, $d_2$ is the distance between the anchor and first negative, and $d_3$ is the distance between the two negatives. We sample these images using the metadata available in our datasets. For instance, every anchor image is tagged with a corresponding target image which will be our positive. Given these two, we sample a large number of negatives from the remaining pool and pick two of them at each step for computing the loss above. 

Our overall loss function is:

\begin{equation}
    L=\lambda \mathcal{L}_\text{MMGrad} + (1-\lambda) \mathcal{L}_\text{quad}
    \label{eq:overallLoss}
\end{equation}

The first term in Equation~\ref{eq:overallLoss} encourages the model to pay attention to the correct local regions when generating the feature vectors whereas the second term ensures these features are discriminative. 

\section{Experiments and Results}
\label{sec:exp_and_res}

We evaluate the performance of our model on a real-world dataset called CIRR (Composed Image Retrieval on Real-life images)~\cite{cirplant}, which is an open-domain dataset for composed image retrieval. It consists of multiple subsets of images (subset size = 6), that are visually ad semantically similar. The modification texts are annotated such that they describe fine differences between images in the same subset. Following~\cite{cirplant}, we report results using two metrics: $Recall@K$ and $Recall_{subset}$. $Recall_{subset}$ is an extension to the standard $Recall$ metric that is based on the unique design of this dataset. $Recall_{subset}$ is computed by ranking images that belong to the related subset. There are two advantages with this metric: (a) It is unaffected false-negative samples, and (b)it evaluates the reasoning ability of the model to capture fine-grained image-text modifications~\cite{cirplant}. We also evaluate our model on Fashion-IQ~\cite{fashioniq}, a fashion dataset with three sub-categories: \textit{Dress, Shirt, Toptee}. We report the standard evaluation metrics followed previously for each sub-category - $Recall@10, Recall@50$, and the average of these two recall values.

\begin{table*}
\begin{center}
\small
\caption{Quantitative results to demonstrate the impact of our proposed MMGrad attention loss. Note: Best numbers are in red.}
\label{tab:withWithoutAttentionNumbers}
\scalebox{1.1}{
\begin{tabular}{lllllllll}
\toprule

Method & \multicolumn{4}{c}{$Recall@K$} & \multicolumn{3}{c}{$Recall_{subset}@K$} & $(R@5 + R_{subset}@1)/2$ \\
        & K=1 & K=5 & K=10 & K=50 & K=1 & K=2 & K=3 & \\
        \hline
        \hline
        Without attention loss & 14.15 & 37.06 & 50.1 & 77.83 & 51.95 & 74.22 & 86.41 & 44.51 \\
        With attention loss & \textbf{\textcolor{red}{19.11}} & \textbf{\textcolor{red}{45.64}} & \textbf{\textcolor{red}{59.81}} & \textbf{\textcolor{red}{86.89}} & \textbf{\textcolor{red}{58.68}} & \textbf{\textcolor{red}{79.90}} & \textbf{\textcolor{red}{90.87}} & \textbf{\textcolor{red}{52.16}} \\
 \midrule

\end{tabular}}
\end{center}
\vspace{-10pt}
\end{table*}

\begin{figure}[th]
    \centering
    \includegraphics[width=\linewidth]{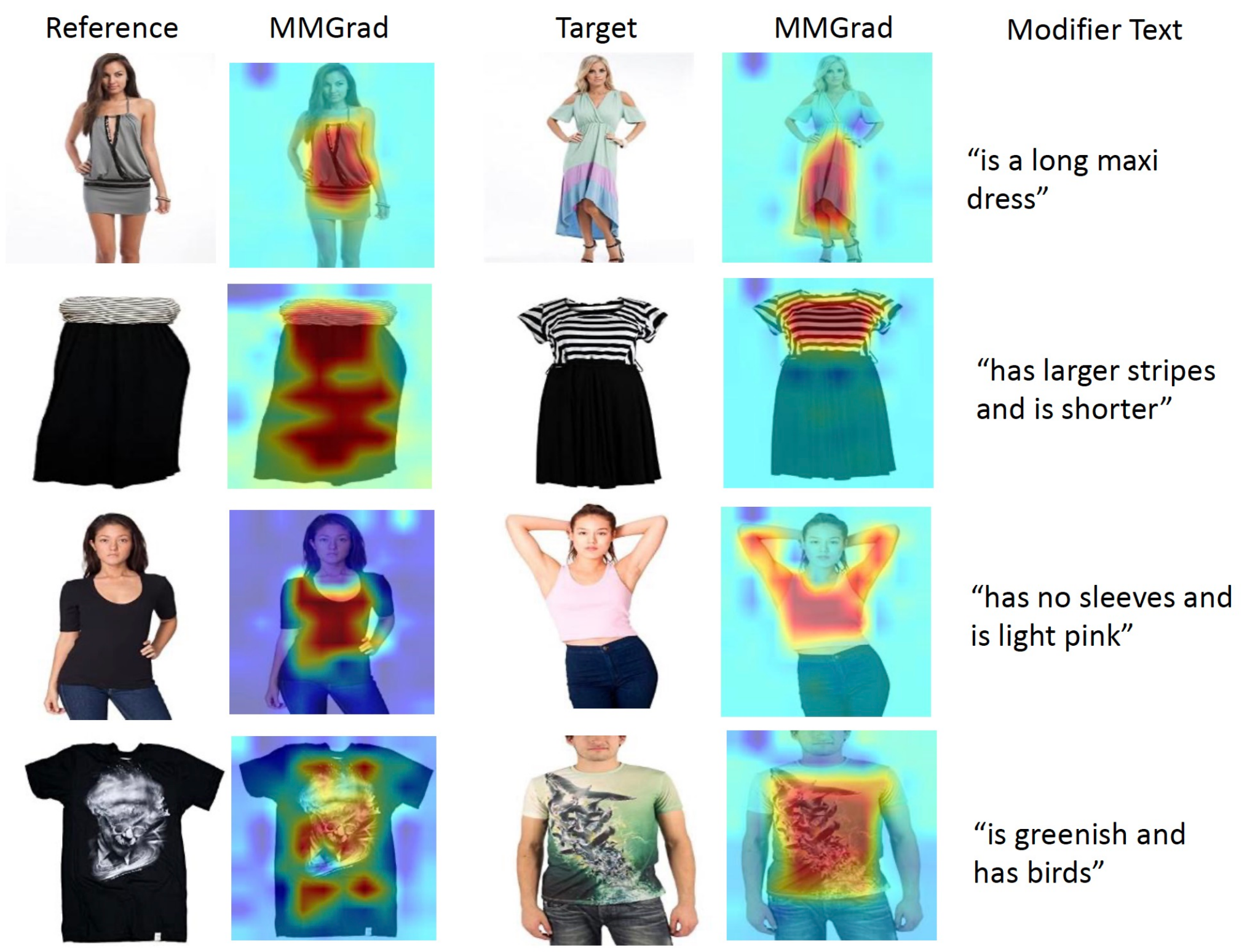}
    \caption{Our attention maps on FashionIQ data.}
    \label{fig:qualResultsFashionIQ}
\end{figure}

\begin{table*}
\begin{center}
\small
\caption{Quantitative comparison with the existing state of the art on the CIRR dataset. Note: Best numbers are in red}
\label{tab:comparison_with_baselines_cirr}
\scalebox{1.0}{
\begin{tabular}{lllllllll}
\toprule

Method & \multicolumn{4}{c}{$Recall@K$} & \multicolumn{3}{c}{$Recall_{subset}@K$} & $(R@5 + R_{subset}@1)/2$ \\
        & K=1 & K=5 & K=10 & K=50 & K=1 & K=2 & K=3 & \\
        \hline
        \hline
        TIRG~\cite{tirg} & 14.61 & \textbf{\textcolor{red}{48.37}} & \textbf{\textcolor{red}{64.08}} & \textbf{\textcolor{red}{90.03}} & 22.67 & 44.97 & 65.14 & 35.52 \\
        TIRG+last Conv~\cite{tirg} & 11.04 & 35.68 & 51.27 & 83.29 & 23.82 & 45.65 & 64.55 & 29.75 \\
        MAAF~\cite{maaf} & 10.31 & 33.03 & 48.30 & 80.06 & 21.05 & 41.81 & 61.60 & 27.04 \\
        MAAF+BERT~\cite{maaf} & 10.12 & 33.10 & 48.01 & 80.57 & 22.04 & 42.41 & 62.14 & 27.57 \\
        MAAF-IT~\cite{maaf} & 9.90 & 32.86 & 48.83 & 80.27 & 21.17 & 42.04 & 60.91 & 27.02 \\
        MAAF-RP~\cite{maaf} & 10.22 & 33.32 & 48.68 & 81.84 & 21.41 & 42.17 & 61.60 & 27.37 \\
        CIRPLANT~\cite{cirplant} & 15.18 & 43.36 & 60.48 & 87.64 & 33.81 & 56.99 & 75.40 & 38.59 \\
        ARTEMIS~\cite{artemis} & 16.96 & 46.10 & 61.31 & 87.73 &39.99 &62.2 & 75.67 & 43.05 \\
         Proposed & \textbf{\textcolor{red}{19.11}} & 45.64 & 59.95 & 86.89 & \textbf{\textcolor{red}{58.68}} & \textbf{\textcolor{red}{79.90}} & \textbf{\textcolor{red}{90.87}} & \textbf{\textcolor{red}{52.16}} \\

 \midrule

\end{tabular}}
\end{center}
\vspace{-20pt}
\end{table*}

\begin{table*}
\begin{center}
\small
\caption{Quantitative comparison with the existing state of the art on the FashionIQ dataset.  $^*$ are reults reported by \cite{cirplant}.}
\label{tab:comparison_with_baselines_fashionIQ}
\scalebox{0.9}{
\begin{tabular}{llllllllll}
\toprule

Method & \multicolumn{2}{c}{Dress@K} & \multicolumn{2}{c}{Shirt@K} & \multicolumn{2}{c}{Toptee@K} & \multicolumn{2}{c}{Avg} 
& (R@10 + R@50)/2 \\
       & R@10 & R@50 & R@10 & R@50 & R@10 & R@50 & R@10 & R@50 \\
       \hline
       \hline
       Image Only$^*$ & 4.20 & 13.29 & 4.51 & 14.47 & 4.13 & 14.30 & 4.28 & 14.20 & 9.15 \\
       Image+Text Concatenation$^*$ & 10.52 & 28.98 & 13.44 & 34.60 & 11.36 & 30.42 & 11.77 & 31.33 & 21.55 \\
       MRN~\cite{mrn} & 12.32 & 32.18 & 15.88 & 34.33 & 18.11 & 36.33 & 15.44 & 34.28 & 24.86 \\
       TIRG~\cite{tirg} & 14.87 & 34.66 & 18.26 & 37.89 & 19.08 & 39.62 & 17.40 & 37.39 & 27.40 \\
       MAAF~\cite{maaf} & 23.8 & 48.6 & 21.3 & 44.2 & 27.9 & 53.6 & 24.3 & 48.8 & 36.6 \\
       CIRPLANT~\cite{cirplant} & 14.38 &  34.66 & 13.64 & 33.56 & 16.44 & 38.34 & 14.82 & 35.52 & 25.17 \\
        Proposed & 14.58 & 34.90 & 12.17 & 31.01 & 14.13 & 35.65 & 13.63 & 33.85 & 23.74 \\
\bottomrule\hline

\midrule
\end{tabular}}
\end{center}
\vspace{-20pt}
\end{table*}

 \begin{figure}[th]
    \centering
    \includegraphics[width=0.9\linewidth]{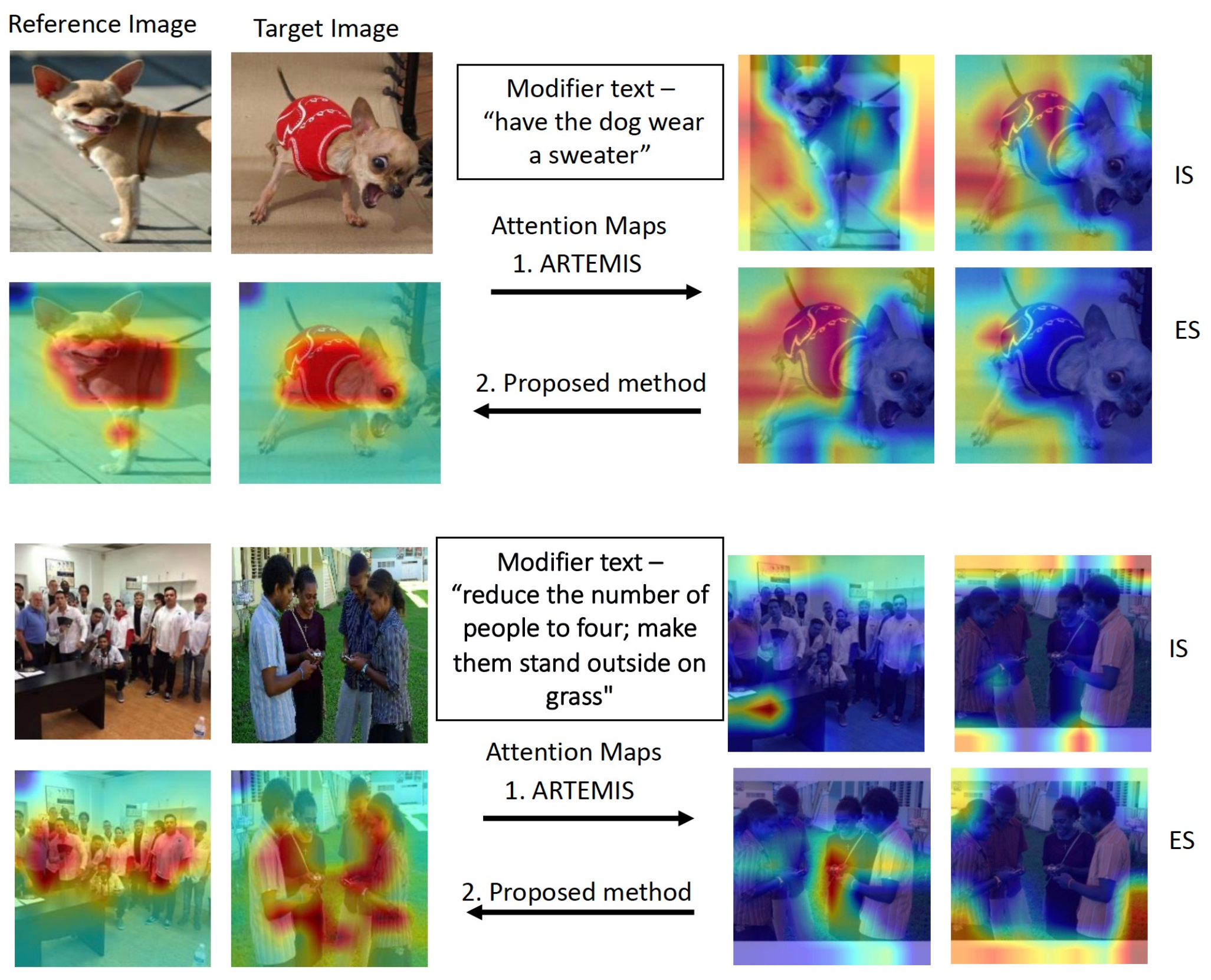}
    \caption{Our attention maps vs. ARTEMIS. IS: visual cues shared between reference and target, ES: regions relevant to the modifier.}
    \label{fig:artemisVsMMGrad}
\end{figure}

\subsection{Qualitative Results}
We begin by showing and discussing qualitative results of our proposed method. Fig.~\ref{fig:qualResults} shows retrievals obtained with our proposed method on six samples, where one can note in all cases our method is able to not only retrieve the correct target image as the rank-1 match but also highlight the right local regions with our MMGrad attention maps. For instance, in the second example (top row, right), our method is able to correctly capture the breed of the dog and consistently cross-referenced it with the input modifier text. Similarly in the first example, our method has not only correctly identified the breed of the dog and retained only one of them from the reference, but has also captured the fine-grained details of a \textit{woman hugging it}. Similar behavior can be observed in other examples as well, including in the (second-row, left) example, where one can note the proposed model is able to correctly capture the \textit{onion rings} and \textit{blue-hue background} from the text and map it to the local regions in the image. In case of the fourth example (second-row, right) our model is able to correctly identify a group of people in the reference image and understand that the number of people has to reduce, along with a change in background, which is correctly captured in the attention map of the retrieved image. It is also interesting to note how our model is able to highlight the dog in the reference image, while correctly highlighting both the panda as well as walls (instead of fence) in the target image (third row, left). 

In Fig~\ref{fig:withWithoutAttention}, we compare the attention map obtained from our model trained separately with and without the proposed MMGrad loss on two samples taken from the validation set of the CIRR dataset. One can note from the results the model with our proposed loss correctly retrieves the ground truth image as the top-1 ranked match, whereas the variant trained without the attention loss is unable to retrieve the correct target image as the rank-1 result. More crucially, from the attention maps, one can note the result with the model with attention loss is focused on the correct regions of interest whereas this is not the case with the variant without the attention loss. For instance, with the first example (on the left), the model with our loss is able to highlight the head of the dog as the region of interest whereas in the second example (on the right), it is able to highlight the group of monkeys on grass correctly. 

\subsection{Quantitative results}
In this section, we report the quantitative performance of our proposed method and compare it with existing state-of-the-art models. We first begin with an ablation study to quantitatively demonstrate the impact of our proposed MMGrad loss of Equation~\ref{eq:mmgradLoss} to further substantiate our notes from Figure~\ref{fig:withWithoutAttention}. In Table~\ref{tab:withWithoutAttentionNumbers}, we show results of two variants of our model- one trained with Equation~\ref{eq:mmgradLoss} and one without. One can note that with Equation~\ref{eq:mmgradLoss}, the model gives a significant improvement across both metrics with $Recall@1$ gains of $\approx 5\%$ and $Recall_{subset}@1$ gains of $\approx 6\%$ over the without-attention-loss version, providing additional evidence to show the utility of highlighting the local regions of interest with our attention maps and using them during training. We next conduct experiments on the test split of the CIRR dataset, use their evaluation server \cite{cirplant} to compute $Recall@K$ and $Recall_{subset}@K$ numbers, and compare results of our model to existing state-of-the-art methods in Table~\ref{tab:comparison_with_baselines_cirr}. We make several observations. First, on the overall averaged metric across both $Recall@5$ and $Recall_{subset}@1$, our proposed method outperforms all the reported methods with a substantial performance gain of $\approx 13\%$. Next, our model also outperforms Cirplant \cite{cirplant}
by a margin of $\approx 4\%$ on $Recall@1$ and a much more substantial $\approx 25\%$ on $Recall_{subset}@1$. All these results demonstrate the need for localization during feature learning which our model achieves with the proposed MMGrad attention loss. In Figure~\ref{fig:artemisVsMMGrad}, we also compare the attention maps and note better localization with our model.

We also conduct experiments on the Fashion IQ dataset. We compare the results of our model with state-of-the-art methods like~\cite{maaf,mrn,tirg,cirplant} in Table~\ref{tab:comparison_with_baselines_fashionIQ}.  A key point to note here is that the images in the Fashion-IQ dataset significantly vary from those in the CIRR dataset. The CIRR dataset contains real-life images. Real-life images depict complex scenarios and the modifier texts correspond to different types of changes. Our model focuses on tackling this type of challenge, while many of the baselines reported in Table~\ref{tab:comparison_with_baselines_fashionIQ} have built models for a specific type of dataset- Fashion data that operates on attribute changes. It is also clearly understood from the results - MAAF~\cite{maaf} achieves the best results on the FashionIQ~\cite{fashioniq} dataset, while it performs poorly when evaluated on the CIRR~\cite{cirplant} dataset. This could potentially mean that the models might have been overly-adapted to the Fashion domain specific images that are usually much less complex than open-domain real-life images. Finally, we also show attention map visualizations on some samples in Figure~\ref{fig:qualResultsFashionIQ} where our method is able to correctly localize the regions of change, e.g., top and sleeves in row 3 and shirt in row 4. We would like to emphasize that in addition to the competitive performance in Table~\ref{tab:comparison_with_baselines_fashionIQ}, our method is able to generate attention maps like in Figure~\ref{fig:qualResultsFashionIQ} that help explain the retrieval results. Such an attention generation mechanism does not exist in the baselines in Table~\ref{tab:comparison_with_baselines_fashionIQ}.

\section{Summary}
In this paper, we considered the problem of composed image retrieval where one is given an input image and a modification text and the task is to retrieve images from a database that semantically satisfy the combined image-text query. We observed that this problem is very local in nature and such local regions have to be highlighted in the learned feature space when doing matching and retrieval. We noted that existing methods instead use global features, leading to a gap in the current literature which we address by learning local features. We realized this by means of a new multi-modal gradient attention (MMGrad) technique that can compute attention maps on images conditioned on input text. We showed how MMGrad can be used to highlight local regions in images that correspond to modification texts in the query and demonstrated it can be computed in a fully differentiable fashion, leading to the end-to-end training of Siamese transformer models with our new attention-based learning objective. We conducted extensive experiments on standard benchmark datasets and showed both qualitative improvements with our MMGrad attention maps as well as competitive quantitative performance.

{\small
\bibliographystyle{ieee_fullname}
\bibliography{egbib}
}
\clearpage

\appendix
\section{Appendix}
\subsection{Implementation Details}
We use the Vision Transformer (ViT) \cite{vit} model as the image encoder $\mathbf{E}_\text{img}$ and the Visual-BERT \cite{visualbert} model as the vision-language transformer $\mathbf{E}_\text{img-text}$.We use ViT's last hidden state and pass it through a Linear Layer to obtain visual embeddings ($f$). We train our model end-to-end using the Siamese training strategy (weight sharing between the two image encoder branches) using the AdamW \cite{loshchilov2017decoupled} optimizer with an initial learning rate of $10^{-5}$ and a batch size of 32 for 350 epochs.

\begin{figure*}[b]
    \centering
    \includegraphics[width=0.85\linewidth]{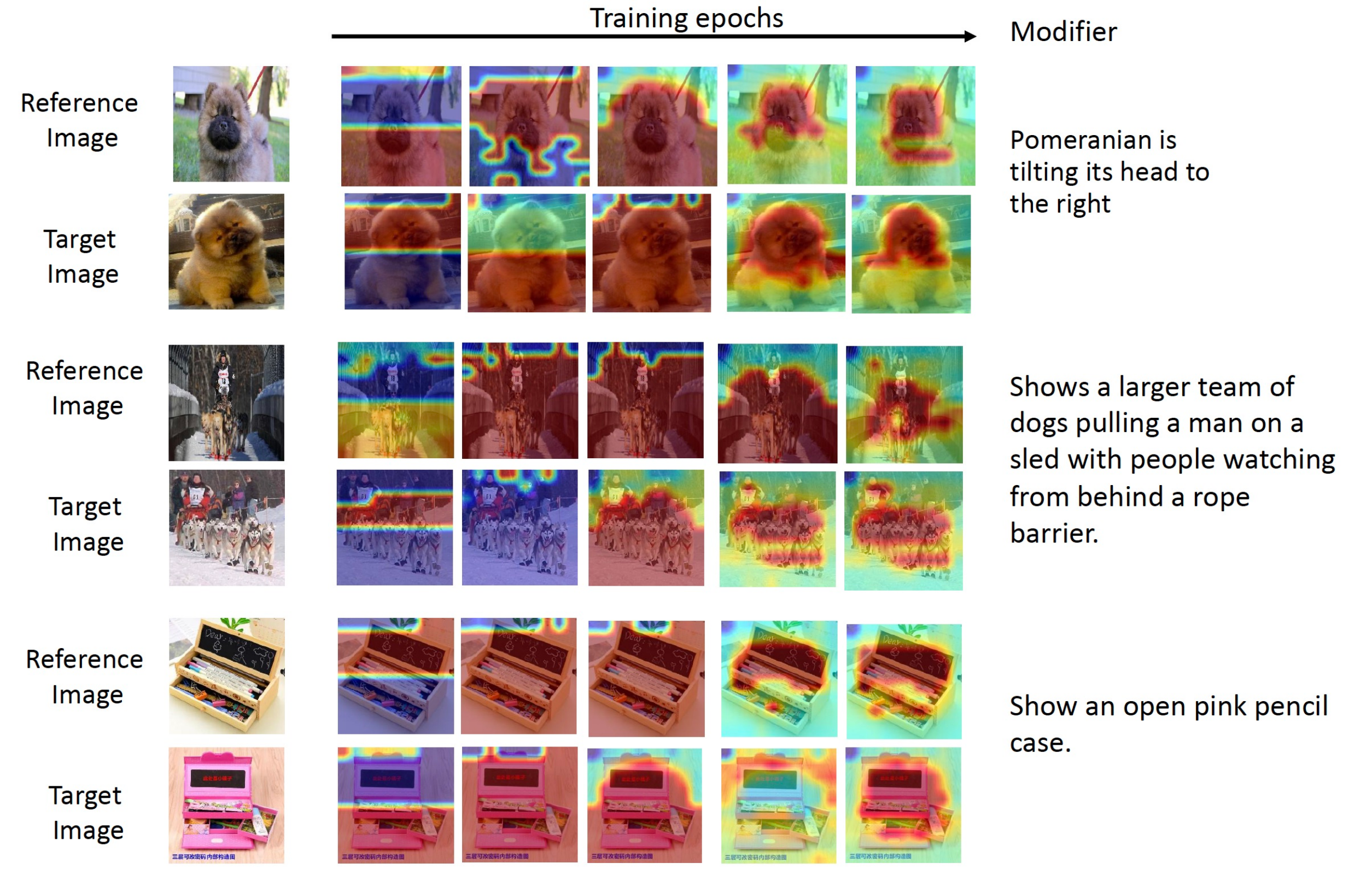}
    \caption{CIRR dataset - Evolution of attention maps over training epochs. }
    \label{fig:evolution}
\end{figure*}

\begin{figure*}
    \centering
    \includegraphics[width=0.85\linewidth]{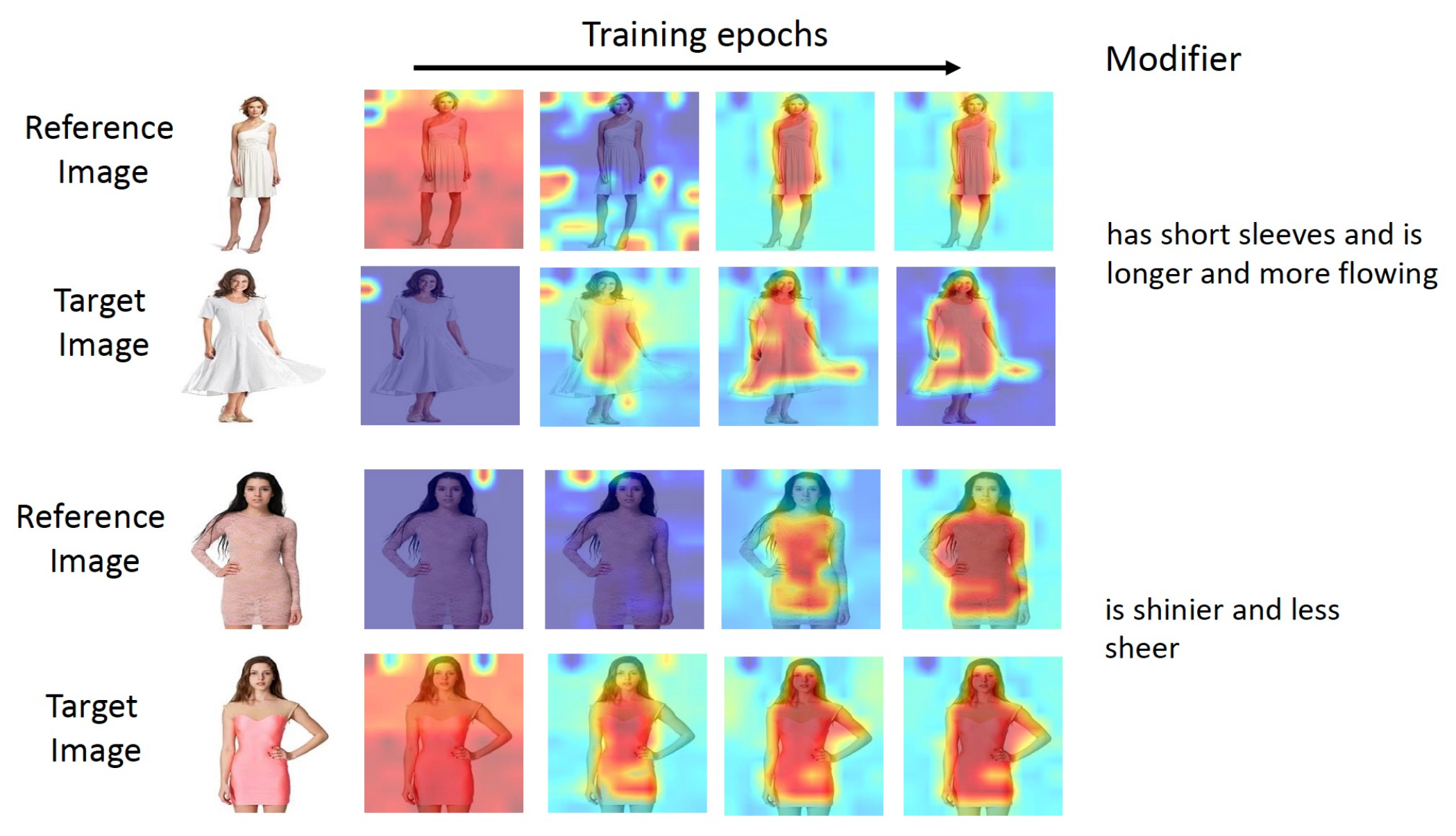}
    \caption{FashionIQ dataset - Evolution of attention maps over training epochs. }
    \label{fig:evolution_fashionIQ}
\end{figure*}

\subsection{Evolution of Attention Maps Over Epochs}
In Figure~\ref{fig:evolution}, we show more examples of attention maps from the training data of CIRR dataset. In Figure~\ref{fig:evolution_fashionIQ}, we show examples of attention maps from the training data of FashionIQ dataset. As with results in Figure 3 in the main paper, one can note the localization of the correct regions improves over training epochs with our proposed MMGrad loss.

\begin{figure*}
    \centering
    \includegraphics[width=0.65\linewidth]{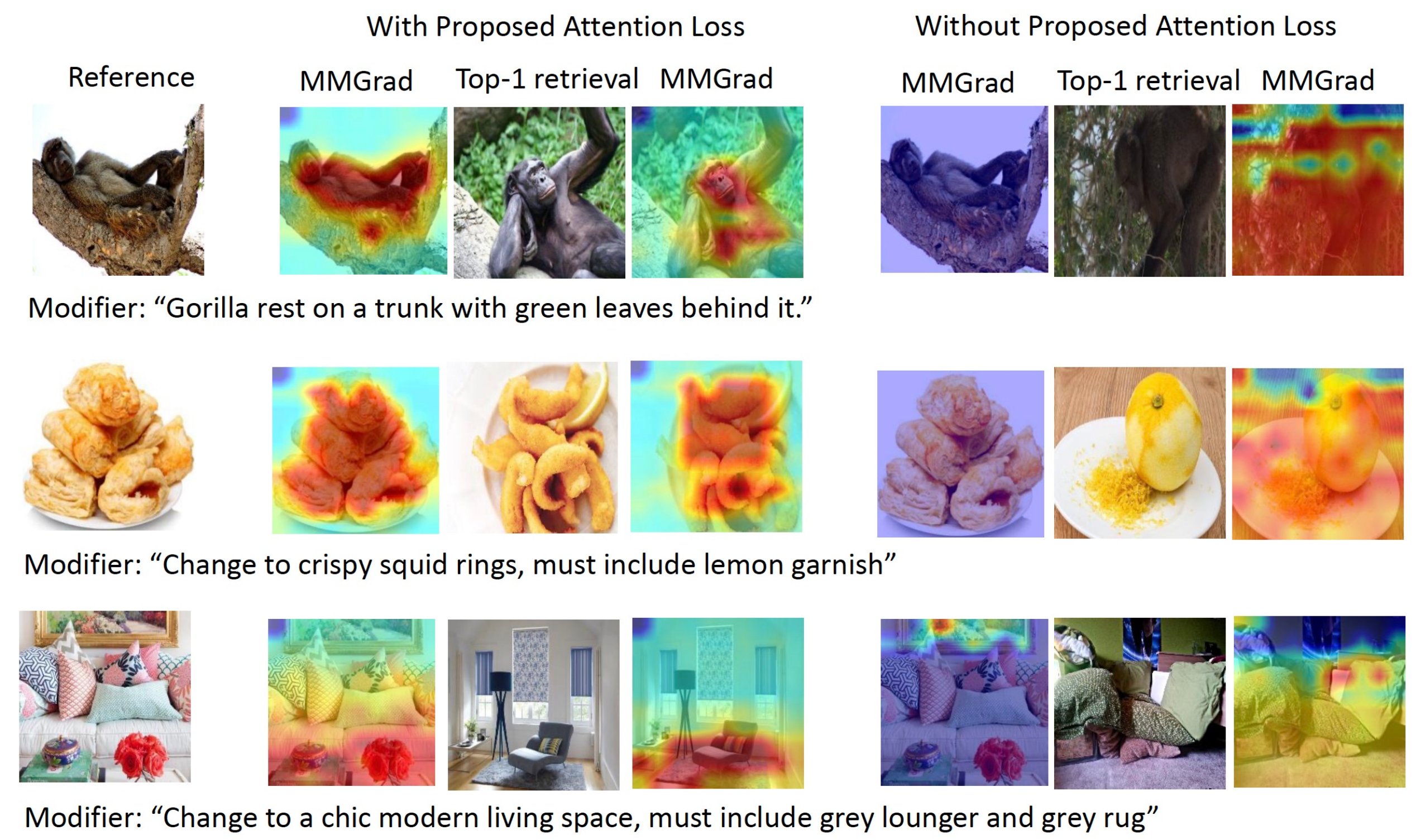}
    \caption{Comparison of the retrievals and the corresponding attention maps obtained from two methods - with the proposed attention loss and without the proposed attention loss.}
    \label{fig:comparison}
\end{figure*}
\begin{figure*}
    \centering
    \includegraphics[width=0.6\linewidth]{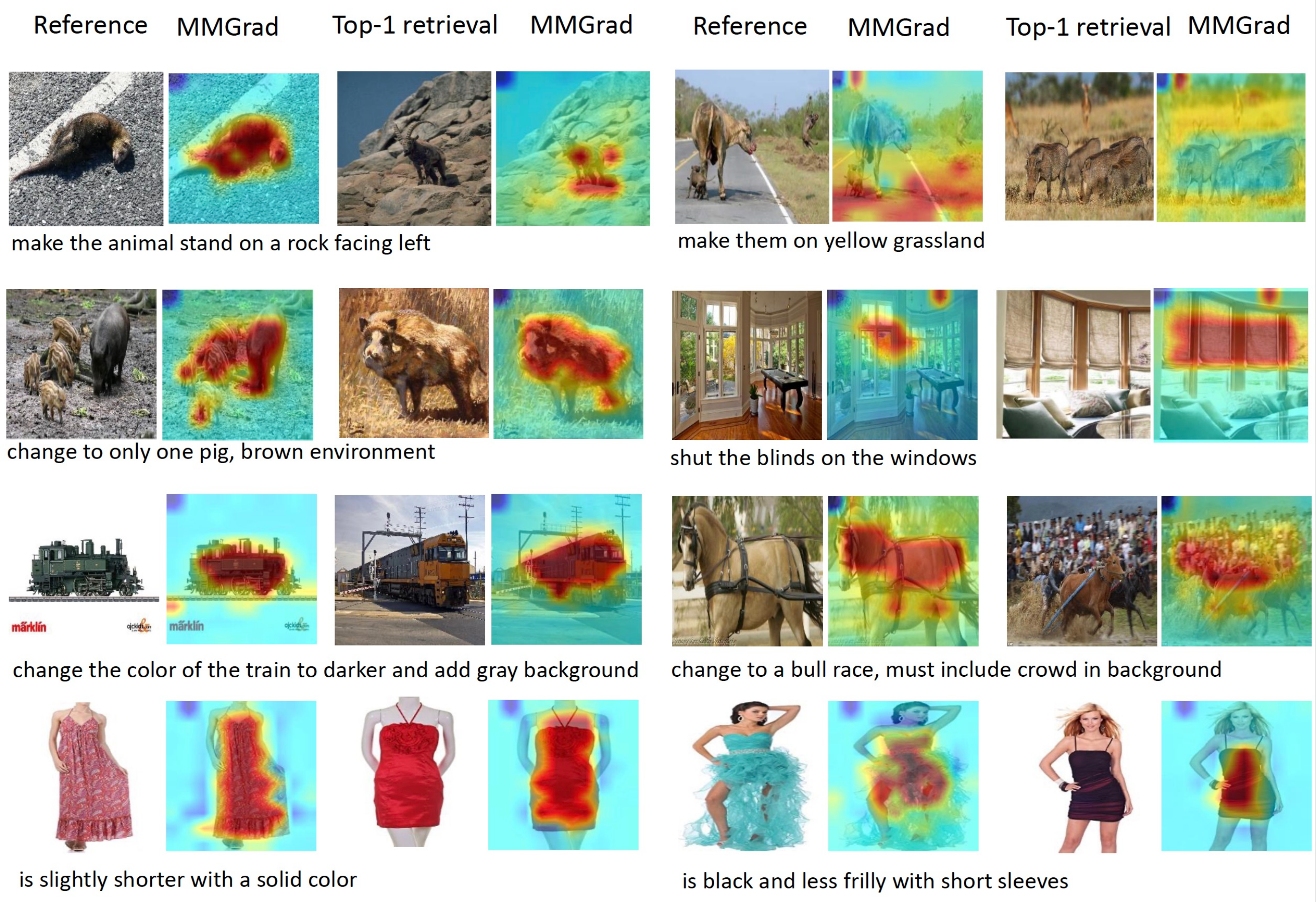}
    \caption{Results on six samples from the test set of the CIRR and two samples from the test set of Fashion IQ dataset using the proposed multi-modal gradient attention based approach.}
    \label{fig:retrievals}
\end{figure*}
\begin{figure*}
    \centering
    \includegraphics[width=0.65\linewidth]{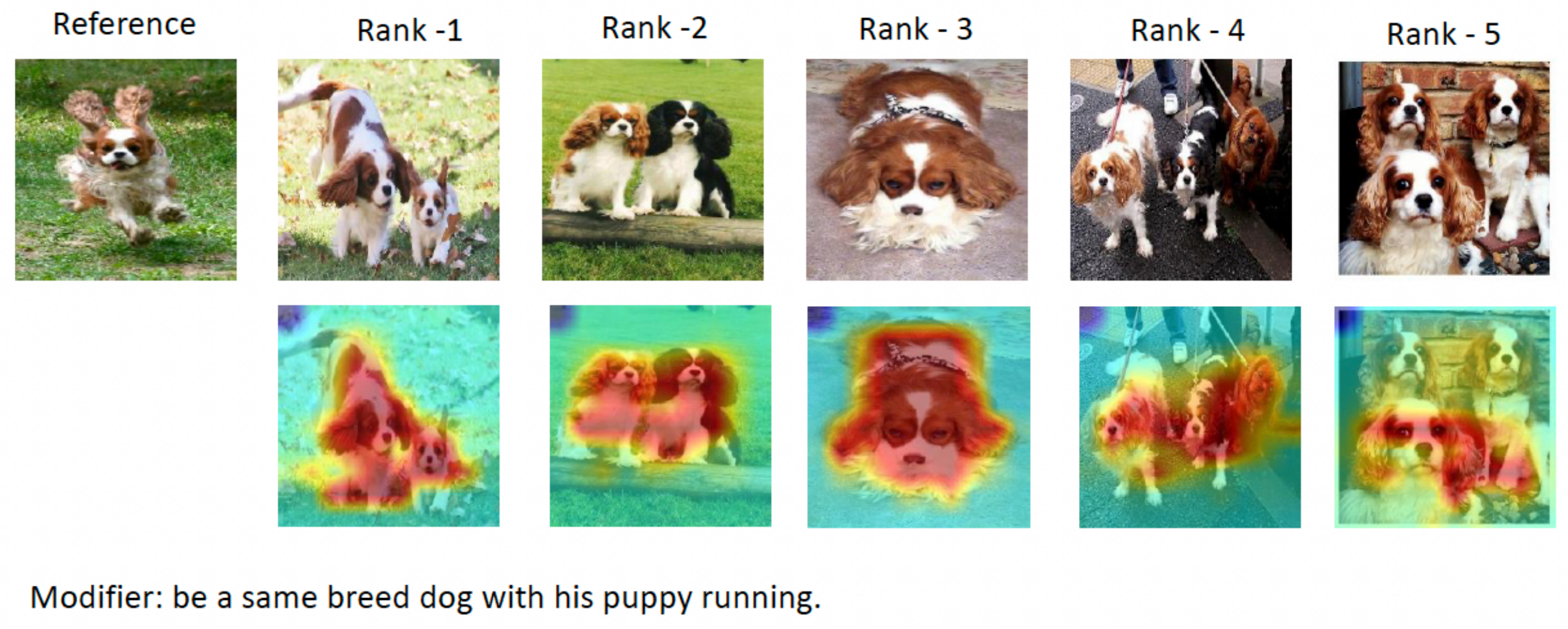}
    \caption{Results on a sample from the validation data of the CIRR dataset.}
    \label{fig:subset}
\end{figure*}

\subsection{Comparison of Results With and Without $L_{MMGrad}$}
In Figure~\ref{fig:comparison}, we show the rank-1 retrieval obtained from two methods - Left: with proposed attention loss; Right: without attention loss. These examples are from the validation set of the CIRR dataset. We note that the rank-1 retrieval obtained from the $L_{MMGrad}$-based proposed model is correct whereas the model without attention loss predicts incorrectly due to the incorrect localization of modifier regions. It is clear from the attention maps that, in the absence of $L_{MMgrad}$, the global features fail to capture local regions of importance, and have attention maps spread across the entire image, thus only capturing global information.

\subsection{Results on the CIRR and Fashion IQ Dataset}
In Figure~\ref{fig:retrievals}, we show the rank-1 retrieval and the corresponding attention map obtained by our proposed MMGrad on test samples of the CIRR and the FashionIQ dataset.

In Figure~\ref{fig:subset}, we show the ranking of images from the subset of visually similar images. As mentioned in the main paper, every reference image in the dataset has a subset of related target images. The rank-1 retrieval of our model is indeed the ground-truth image. This highlights the reasoning ability of our approach to capture fine-grained image-text modifications.

\end{document}